\documentclass[letterpaper, 10 pt, conference]{ieeeconf}  

\IEEEoverridecommandlockouts                              

\usepackage{graphicx}
\usepackage{booktabs}
\usepackage{multirow}
\usepackage{xspace}
\usepackage{arydshln}
\usepackage{colortbl}
\usepackage{xcolor}
\usepackage{caption}
\usepackage{amsmath,amssymb,mathtools}
\usepackage{bm} 
\usepackage{kotex}
\usepackage{placeins}
\usepackage{tikz}
\usetikzlibrary{fit,calc}
\usepackage{tikz}
\usetikzlibrary{tikzmark}
\usepackage{pifont}
\usepackage{flushend}

\usepackage{algorithmic}
\usepackage{algorithm}

\makeatletter
\DeclareRobustCommand\onedot{\futurelet\@let@token\@onedot}
\def\@onedot{\ifx\@let@token.\else.\null\fi\xspace}

\def\eg{\emph{e.g}\onedot} 
\def\ie{\emph{i.e}\onedot} 
 
 \def\vs{\emph{vs}\onedot}

\makeatother

\newcommand{\methodfull}{\mbox{\textbf{B}ridging \textbf{IN}stant and \textbf{DE}liberative \textbf{R}easoning}\xspace}
\newcommand{\method}{\mbox{\textsc{BINDER}}\xspace}

\definecolor{drmBG}{RGB}{235,243,255}   
\definecolor{irmBG}{RGB}{255,244,229}   
\definecolor{mainAccent}{RGB}{0,0,0} 
\definecolor{drmAccent}{RGB}{30,70,160} 
\definecolor{irmAccent}{RGB}{180,90,0}  
\definecolor{mainBG}{RGB}{238,245,255} 

\usepackage{hyperref}
\hypersetup{
    colorlinks=true,
    linkcolor=blue,
    citecolor=blue,
}

\title{\LARGE \bf \method: Instantly Adaptive Mobile Manipulation with Open-Vocabulary Commands}

\author{Seongwon Cho$^{*}$, Daechul Ahn$^{*}$, Donghyun Shin, Hyeonbeom Choi, San Kim, Jonghyun Choi$^{\dagger}$
\thanks{All authors are with Seoul National University, Seoul 08826, Republic of Korea. $^{*}$ indicates equal contribution. $^{\dagger}$JC is with ECE, ASRI and IPAI in SNU and a corresponding author ({\tt\footnotesize jonghyunchoi@snu.ac.kr}).}%
\\[0.1cm]
\texttt{\href{https://seongwon980.github.io/projects/binder}{\textbf{\texttt{https://seongwon980.github.io/projects/binder}}}}%
}

\begin{document}

\maketitle
\thispagestyle{empty}
\pagestyle{empty}

\begin{abstract}
Open-vocabulary mobile manipulation (OVMM) requires robots to follow language instructions, navigate, and manipulate while updating their world representation as the environment changes dynamically.
However, most prior works update their world representation only at discrete milestones, such as waypoints or the end of an action step.
Such sparse updates leave robots with limited awareness between updates, causing missed objects, delayed error detection, and slower replanning.
To address this limitation, we propose \method (\methodfull), a dual-process framework that separates strategic planning from continuous environmental monitoring.
\method combines a Deliberative Response Module (DRM, a multimodal LLM for task planning) with an Instant Response Module (IRM, a Video-LLM for continuous monitoring).
The DRM handles strategic planning through structured 3D scene updates and guides the IRM's focus, while the IRM processes video streams to update memory, proactively adjust actions, and trigger replanning when needed.
This bidirectional coordination ensures continuous awareness without costly updates, enabling reliable and robust operation under dynamic conditions.
We evaluate \method in three real-world environments where objects are moved during execution and show that it achieves substantially higher success rates and efficiency than state-of-the-art baselines, confirming its effectiveness for real-world deployment.
\end{abstract}

\section{Introduction}
\label{sec:intro}
Open-Vocabulary Mobile Manipulation (OVMM) aims to enable robots to navigate unknown environments and manipulate objects based on open-vocabulary instructions~\cite{yenamandra2023homerobot, wu2023tidybot}.
In real-world settings such as homes and offices, robots must cope with dynamic changes like object relocation and human movement, requiring both strategic planning and continuous monitoring.
While prior approaches operated in fixed, pre-scanned environments without considering such changes~\cite{liu2024okrobot,werby2024hierarchical,shafiullah2022clip}, recent approaches incorporate environmental feedback through voxel maps~\cite{liu2024dynamem}, scene graphs~\cite{yan2024dovsg,mohammadi2025more}, and vision--language models for closed-loop reasoning~\cite{comerobot2024}.

\begin{figure}[!t]
    \centering
    \includegraphics[width=\linewidth]{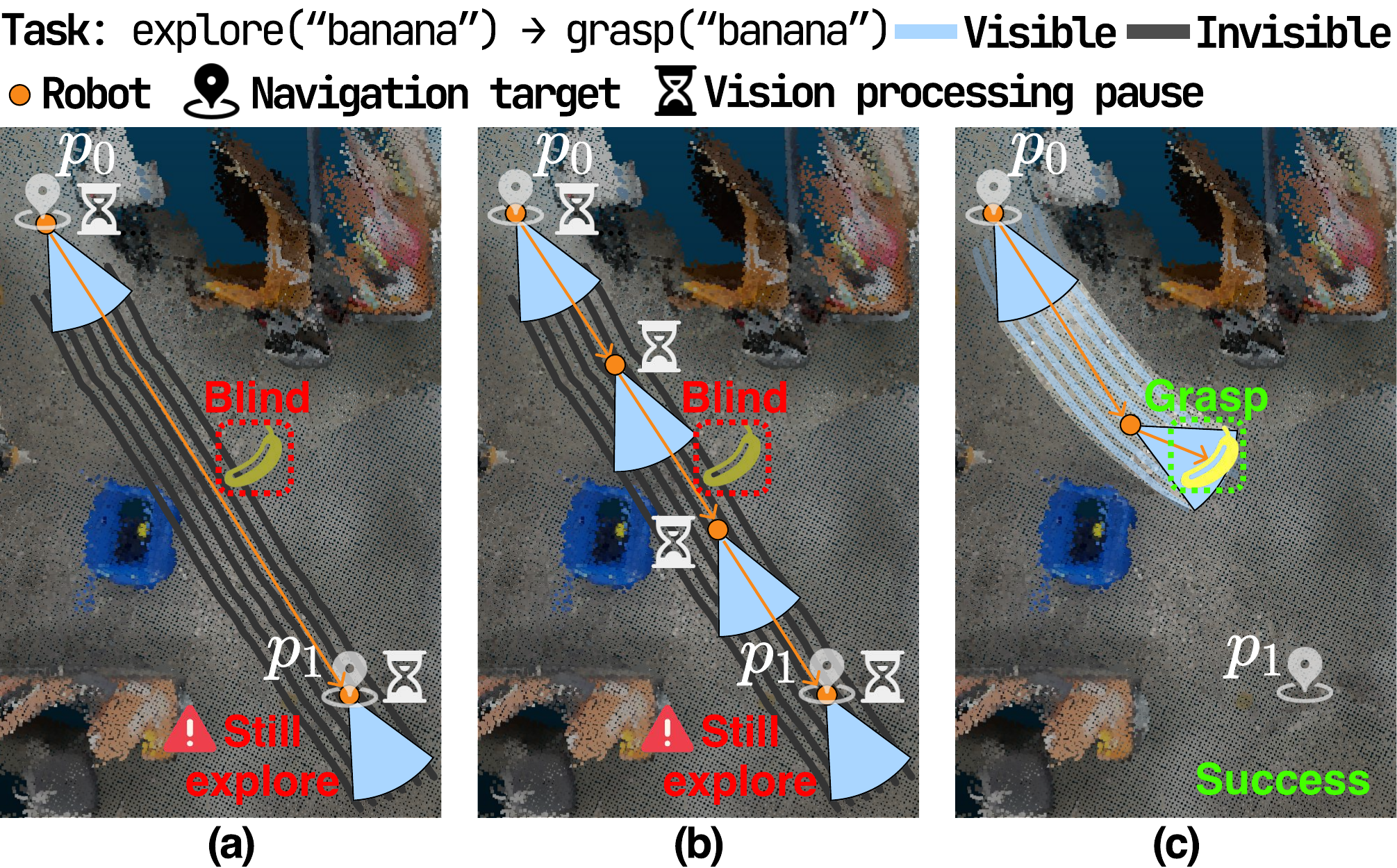}
    \caption{\noindent \textbf{Limitations of existing OVMM approaches and the proposed \method.}
Robots search for a banana while moving from navigation target $p_{0}$ to $p_{1}$.
    \textbf{(a)} Sparse-update approaches refresh perception only at navigation targets, leaving robots unaware during traversal and causing missed objects en-route.
    \textbf{(b)} Frequent updates at intermediate waypoints improve environmental awareness but require repeated vision processing pauses, causing inefficiency and still leaving perception gaps.
    \textbf{(c)} \method maintains continuous visual awareness en-route, enabling opportunistic detections and task execution without extra vision processing pauses.}
    \label{fig:motivation}
\end{figure}


However, these approaches rely on \emph{intermittent scene perception}, leaving robots with limited awareness of environmental changes between updates.
Because 3D semantic reconstruction is computationally expensive, environmental representations---whether 3D voxel maps~\cite{liu2024dynamem,comerobot2024} or scene graphs~\cite{yan2024dovsg,mohammadi2025more,werby2024hierarchical}, or implicit or object-centric maps~\cite{shafiullah2022clip,comerobot2024,qiu2024open}---are refreshed only at \emph{discrete intervals}~\cite{liu2024dynamem,yan2024dovsg,comerobot2024,mohammadi2025more}.
Even approaches employing powerful task planners (\eg, GPT~\cite{yan2024dovsg,comerobot2024}) remain limited by intermittent perception, as their reasoning may rely on outdated scene information.

Consider a robot searching for `banana' as it travels from $p_{0}$ to $p_{1}$, as illustrated in Fig.~\ref{fig:motivation}.
Even if the object lies directly on its path, approaches that update 3D semantic scenes only at navigation targets or after sub-actions (\eg, grasping/placing) can miss this opportunity (Fig.~\ref{fig:motivation}-(a)).
More frequent updates---at intermediate checkpoints~\cite{liu2024dynamem} or during frontier expansions~\cite{honerkamp2024language}---can still miss changes between reconstruction intervals (Fig.~\ref{fig:motivation}-(b)).
This temporal unawareness, inherent to discrete-update approaches, triggers a chain of inefficiencies. Robots might ignore clearly visible objects while exploring, and during manipulation, they can fail to notice minor shifts that escalate into grasp failures, trajectory deviations, collisions, and task breakdowns.

Since 3D semantic scene reconstruction can take tens of seconds to minutes per update depending on scene complexity~\cite{yan2024dovsg,comerobot2024}, robots face an unsatisfactory trade-off: either pause frequently for accurate scene updates---delaying task completion---or continue moving with potentially outdated spatial information---risking critical oversights.
While fast geometric reconstruction algorithms exist~\cite{newcombe2011kinectfusion,tosi2024nerfs}, scaling them to the semantic level remains computationally prohibitive.
To address this computational constraint, we argue that instead of relying solely on monolithic 3D reconstruction, a heterogeneous perception strategy can offer a practical alternative by exploiting the complementary strengths of different sensing modalities: video streams provide continuous semantic awareness and detect salient environmental changes, while 3D reconstruction delivers the precise geometric information essential for OVMM task planning.
By separating semantic monitoring from geometric perception, robots can maintain environmental awareness through video stream analysis while reserving computationally intensive 3D reconstruction for OVMM task planning.

\begin{figure*}[!t]
    \centering
    \includegraphics[width=1\linewidth]{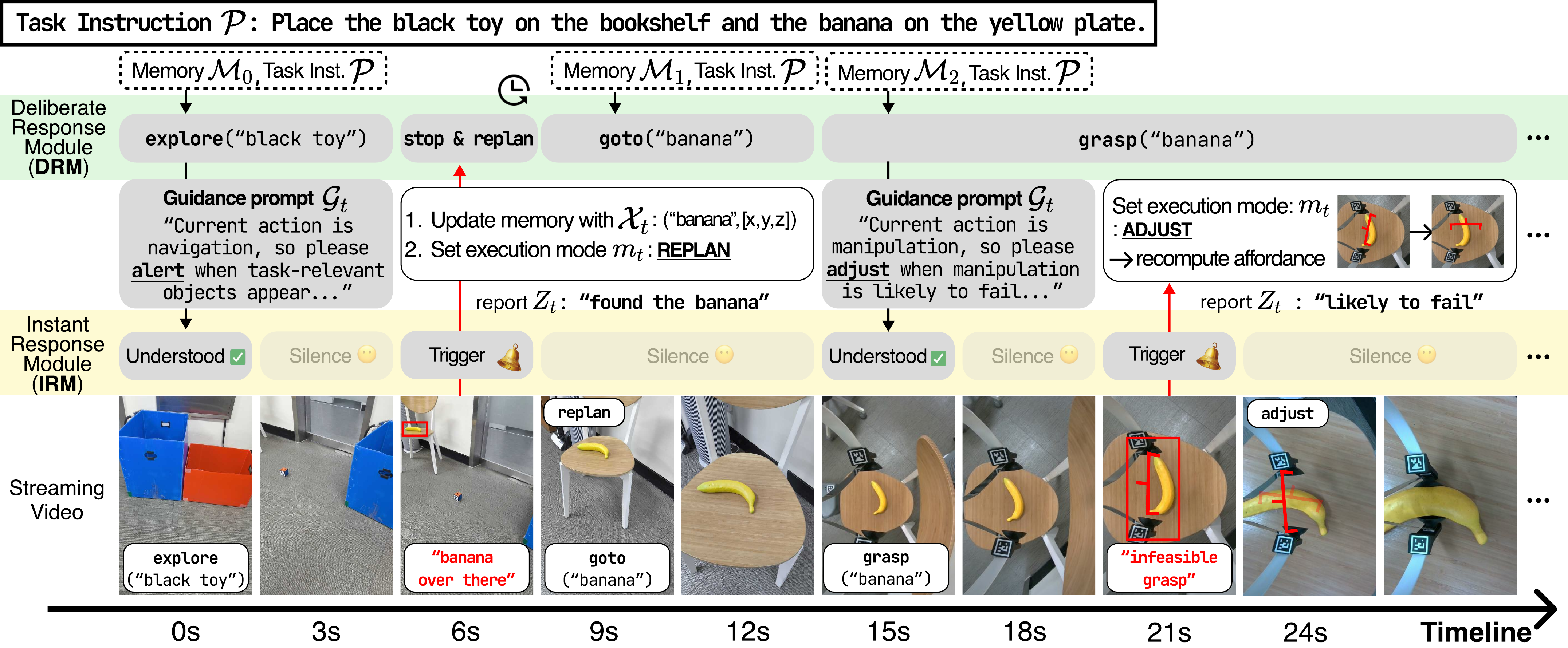}
    \caption{\noindent \textbf{Illustration of dual-process reasoning in \method.} 
    Our proposed framework, BINDER, consists of two modules operating in parallel: \textit{Deliberative Response Module (DRM)} and \textit{Instant Response Module (IRM)}.
    Based on the task instruction (inst.) and memory, DRM issues high-level actions (\eg, \texttt{explore(``black toy'')}) and guides IRM's attention.
    In parallel, IRM continuously monitors the video stream in the background.
    When a task-relevant event occurs---such as opportunistically detecting the task-relevant object (6s) or diagnosing a grasp failure (21s)---IRM immediately generates a report, prompting DRM to replan for navigation or adjust the grasp for manipulation.
    This bidirectional coordination enables both continuous responsiveness and adaptive planning, addressing the intermittent scene perception of prior OVMMs.}
    \label{fig:overview}
\end{figure*}

To this end, we propose \method (\textbf{B}ridging \textbf{IN}stant and \textbf{DE}liberative \textbf{R}easoning), a dual-process framework inspired by cognitive theories~\cite{wason1974dual,kahneman2011thinking} that describe how humans navigate complex environments through fast, automatic monitoring (System 1) and slow, deliberative reasoning (System 2).
Our framework operationalizes this cognitive division through two distinct modules (Fig.~\ref{fig:overview}). The Instant Response Module (IRM) relies on a Video-LLM~\cite{qwen25vl2025} to continuously process video streams, enabling opportunistic interventions during navigation and manipulation. Meanwhile, Deliberative Response Module (DRM) performs strategic planning using 3D semantic scene representations, which update upon navigation targets or when triggered by the IRM.
Furthermore, to enable mutual enhancement between these modules, we propose a bidirectional coordination method.
Specifically, the DRM guides the IRM's monitoring attention based on current task context—whether navigating, searching, or manipulating—ensuring situation-appropriate monitoring, while the IRM provides environmental observations that enable context-aware planning and, when necessary, trigger immediate 3D reconstruction and replanning by the DRM (Sec.~\ref{subsec:dual_process}).
This heterogeneous perception—combining scheduled reconstruction at navigation targets with on-demand analysis from video—addresses the trade-off between temporal awareness and spatial precision that limits monolithic approaches.

We evaluate \method through extensive experiments across three real-world environments featuring diverse dynamic scenarios.
When tested with dynamically appearing/disappearing objects and changing receptacles, \method demonstrates several key capabilities: immediate grasp correction during manipulation, early failure detection through temporal cues, opportunistic replanning when detecting targets mid-navigation, and dynamic task reordering based on environmental changes.
Compared to state-of-the-art baselines~\cite{liu2024okrobot,liu2024dynamem,yan2024dovsg}, our approach shows significant improvements in handling dynamic situations—validating its potential for real-world OVMM deployment.

We summarize our contributions as follows:
\begin{itemize}
\setlength\itemsep{-0.1em}
\item We identify \emph{intermittent scene perception as a limitation of current OVMM systems} and propose \method, a dual-process framework that decouples continuous video monitoring from selective 3D reconstruction.
\item We develop a bidirectional coordination mechanism enabling the IRM to trigger on-demand 3D updates while the DRM guides task-aware monitoring.
\item We demonstrate through real-world experiments that \method effectively handles dynamic scenarios, significantly improving success rates and reducing task completion time compared to state-of-the-art baselines.
\end{itemize}

\section{Related Work}

\subsection{Open Vocabulary Mobile Manipulation}
Open-vocabulary mobile manipulation (OVMM) remains a demanding problem in robotics, as it combines navigation, manipulation, and language understanding over extended horizons, often within environments subject to dynamic change. 
Two main paradigms have emerged for OVMM. One line of work trains end-to-end vision–language–action (VLA) policies on large-scale demonstration corpora and deploys them directly in real-world tasks~\cite{zitkovich2023rt,kim24openvla,black2024pi0,shi2025hirobot}. 
While such models capture rich multimodal correlations and generalize to a wide range of instructions, they suffer from high computational cost and limited scalability to multi-step, long-horizon tasks. 
Moreover, the lack of explicit memory or structured planning makes it difficult to recover from failures or adapt to environmental changes. 

In contrast, modular pipelines decompose perception, grounding, planning, and control into separate components~\cite{yenamandra2023homerobot,liu2024okrobot,liu2024dynamem,yan2024dovsg}. 
By leveraging large language models (LLMs) that synthesize executable action code from language~\cite{liang2022code} and vision-language models (VLMs) for perception and object grounding~\cite{huang2022visual,chen2022open}, modular approaches enable robots to follow diverse natural language instructions. More recent systems such as LOVMM~\cite{tan2025language} combine an LLM for interpreting free-form instructions with VLM-based semantic mapping, enabling language-conditioned navigation and cross-workspace manipulation in household environments.
However, such open-loop pipelines frequently accumulate cross-module errors~\cite{sui-etal-2025-grounding}, degrading performance over long horizons.
Unlike prior OVMM paradigms, we directly address perception intermittency by decoupling continuous video monitoring from selective 3D reconstruction.



\subsection{Closed-Loop Recovery in Robotic Systems}
Recent work incorporates closed-loop recovery to reduce cascading errors.
COME-Robot~\cite{comerobot2024} uses GPT-4V to repeatedly observe the scene, verify task progress, and invoke iterative replanning whenever a subtask is judged to have failed, demonstrating that powerful VLMs can substantially improve robustness through situated reasoning and feedback-driven restoration.
RACER~\cite{dai2025racer} instead learns rich language-guided failure recovery policies: a VLM supervisor provides corrective language feedback, while an actor policy executes visuomotor skills conditioned on these recovery descriptions, yielding strong performance on long-horizon manipulation benchmarks.
Beyond these OVMM-oriented or language-guided systems, related efforts such as language-driven closed-loop grasping with online 6D pose tracking and model-predictive control~\cite{nguyen2025language}, Code-as-Monitor (CaM) for reactive and proactive failure detection via VLM-generated monitors~\cite{zhou2025cam}, and CLOVER’s closed-loop visuomotor control with generative visual plans~\cite{bu2024clover} further highlight the importance of feedback for robust execution.

However, these systems typically assess the results at the level of completed actions or macro steps—checking whether a grasp, placement, or short skill has succeeded—before triggering recovery, which constrains responsiveness when small deviations arise mid-execution.
Our IRM instead issues \textsc{continue}/\textsc{adjust}/\textsc{replan} signals \emph{during} execution, based on continuous video monitoring, enabling fine-grained, timely corrections that prevent minor pose or state errors from compounding into full task failures.

\subsection{Scene Representations for OVMM}
Robust scene representations enable OVMM by preserving object-level semantics and relations for long-horizon reasoning. Graph-based methods fuse multi-view evidence and support scalable queries: ConceptGraphs~\cite{gu2024conceptgraphs} builds an open-vocabulary scene graph; HOV-SG~\cite{werby2024hierarchical} adds a floor–room–object hierarchy for large-scale, multi-floor navigation; and DovSG~\cite{yan2024dovsg} performs local, in-place 3D updates during interaction without full reconstruction.
Voxel/field methods encode language-conditioned 3D maps: CLIP-Fields~\cite{shafiullah2022clip} enables continuous queries via implicit fields; VLMaps~\cite{huang2022visual} grounds features in 3D spatial map for language-driven navigation; and DynaMem~\cite{liu2024dynamem} introduces sparse and efficient updates for long horizons. More recent work~\cite{zhang2025open} further augments open-vocabulary 3D scene graphs with functional relationships and interactive elements, enabling functional reasoning for indoor manipulation. Yet all rely on discrete refreshes, so mid-execution changes can be missed and maps drift. In contrast, our dual-process design maintains continuous awareness and triggers 3D updates.


\section{Approach}
OVMM in dynamic settings demands continuous perception and adaptive planning to handle appearing/relocating objects and to monitor/correct manipulation errors. Yet prior systems use \emph{intermittent scene perception} (limited by compute constraints), leaving robots with limited awareness between discrete updates. \method\ is a dual-process framework that decouples strategic planning from continuous monitoring, delivering strong reasoning with real-time environmental awareness under dynamic conditions.

\begin{figure}[!t]
  \centering
  \includegraphics[width=1\linewidth]{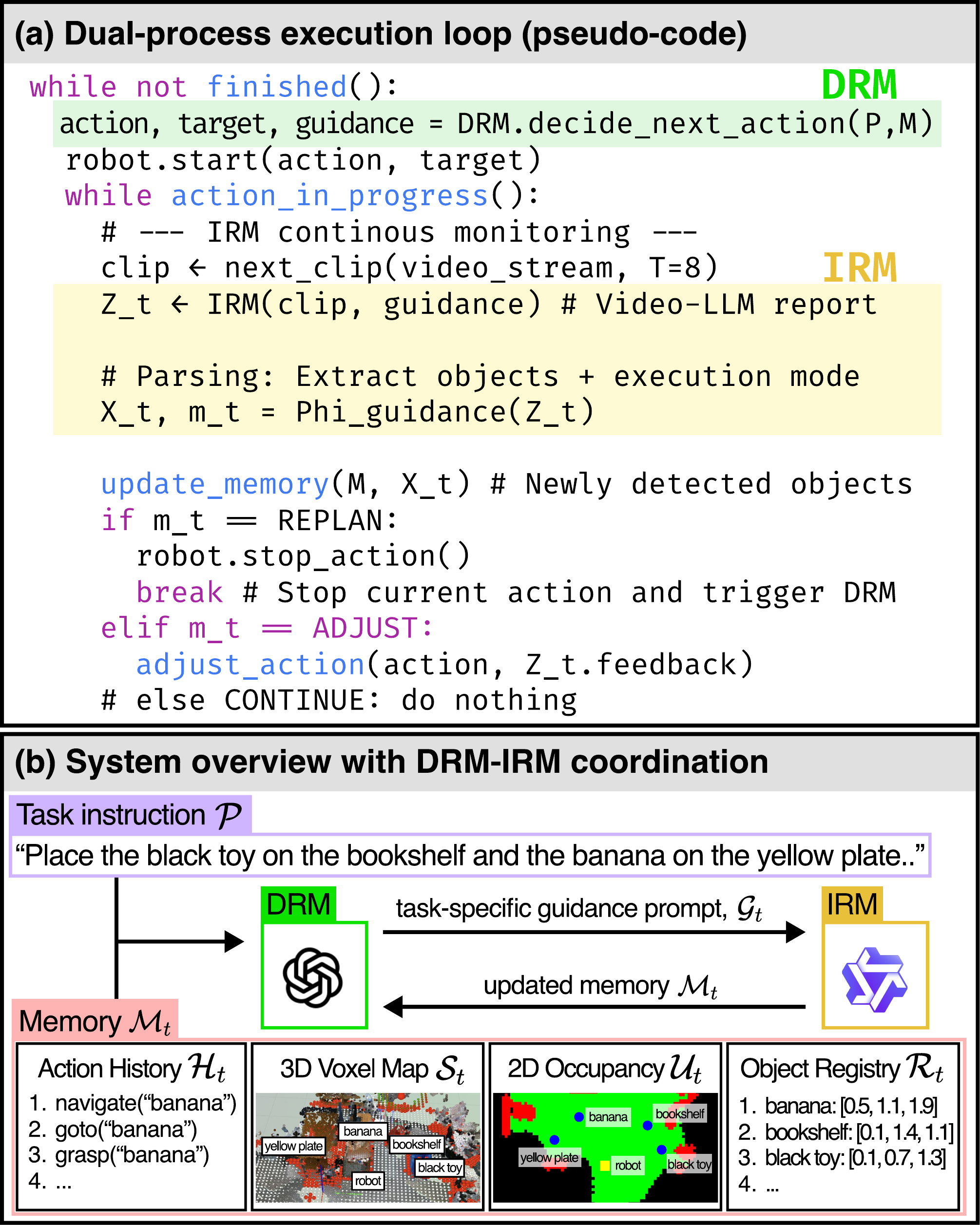}
  \caption{\noindent \textbf{Flowchart of dual-process execution in \method.}
  (a) Pseudocode of the execution loop: the DRM issues high-level actions \emph{and task-specific guidance}, while the IRM continuously monitors video and outputs execution modes (\textsc{continue}/\textsc{adjust}/\textsc{replan}) and object updates that drive local corrections or trigger replanning. 
  (b) System overview: the DRM uses task instructions and memory to generate plans and guidance, while the IRM monitors environmental changes to update memory status and trigger timely replanning under dynamic conditions.}
  \label{fig:flow}
\end{figure}

\subsection{\method: Dual-Process Framework for OVMM}
Existing approaches for OVMM reveal a fundamental limitation: they apply the same computationally expensive 3D semantic scene reconstruction for all perception tasks, creating an unnecessary trade-off between awareness and efficiency. Frequent updates ensure awareness but degrade efficiency, while sparse updates maintain speed but miss critical changes~\cite{liu2024dynamem,yan2024dovsg,comerobot2024}.

\vspace{0.5em}
\noindent \textbf{Dual-process architecture.}
We posit that this trade-off stems from treating all perception tasks as equally demanding: previous approaches apply computationally expensive 3D reconstruction uniformly, without distinguishing between tasks that require geometric precision and those that do not.
While planning tasks necessarily require precise 3D geometry for manipulation and navigation decisions, we argue that monitoring tasks—detecting new objects or environmental changes—can be effectively handled through continuous video analysis, avoiding costly reconstruction overhead during navigation without sacrificing environmental awareness.
This natural division between computationally-intensive planning and relatively lightweight monitoring parallels how humans navigate complex environments—through both fast, automatic monitoring (System 1) and slow, deliberative response (System 2), as described in dual-process theories~\cite{wason1974dual,kahneman2011thinking}.
Inspired by this well-established cognitive architecture, we decouple continuous environmental monitoring from costly periodic 3D reconstruction.

To operationalize this separation, we introduce two specialized modules as illustrated in Fig.~\ref{fig:overview}: the \textbf{Instant Response Module (IRM)} powered by a Video-LLM maintains continuous environmental monitoring through video streams during execution, analogous to System 1's automatic processing; while the \textbf{Deliberative Response Module (DRM)} performs strategic planning using 3D semantic scene representations at navigation targets, mirroring System 2's deliberative reasoning.
This architectural division allows each module to optimize for its primary objective—the IRM for temporal responsiveness, the DRM for spatial precision—enabling both continuous awareness and sophisticated reasoning without the compromises of current monolithic approaches.

\subsection{Dual-Process Modules}
\label{subsec:dual_process}

\vspace{0.5em}
\noindent \textbf{DRM-IRM coordination.}
While the DRM and IRM serve distinct roles, effective OVMM requires coordination between continuous monitoring and strategic planning (Fig.~\ref{fig:overview}).
We achieve this through bidirectional information flow between the modules, as illustrated in Fig.~\ref{fig:flow}. 
The DRM provides task-specific guidance prompt $\mathcal{G}_{t}$ that dynamically reconfigures the IRM's attention—shifting from ``identify task-relevant objects and receptacles'' during exploration to ``monitor gripper-object alignment and placement stability'' during manipulation.
Conversely, the IRM supplies continuous environmental feedback: during exploration, newly detected or relocated objects asynchronously update the object registry $\mathcal{R}_t$ without full map reconstruction; during manipulation, it enables reactive control through immediate local corrections or escalation to the DRM when local adjustments fail.

This bidirectional coordination ensures the system remains both deliberate and responsive. During the execution of an action, the IRM handles transient scene changes and small execution deviations in the background while the robot continues moving, and the robot only stops when the IRM identifies a situation that invalidates the ongoing plan. As a result, \method avoids unnecessary stop-and-update cycles while still reacting promptly to task-relevant changes, mitigating the trade-off between maintaining awareness and the cost of full-scene updates. This coordination also addresses cases where objects move during a 3D reconstruction phase: although such movement can render parts of the reconstructed map outdated, the IRM continues to monitor the video stream and updates the object registry $\mathcal{R}_t$ when an object is detected at a new position, ensuring that subsequent planning uses the most recent location without requiring a full re-reconstruction.

\begin{figure*}[t]
    \centering
    \includegraphics[width=1\linewidth]{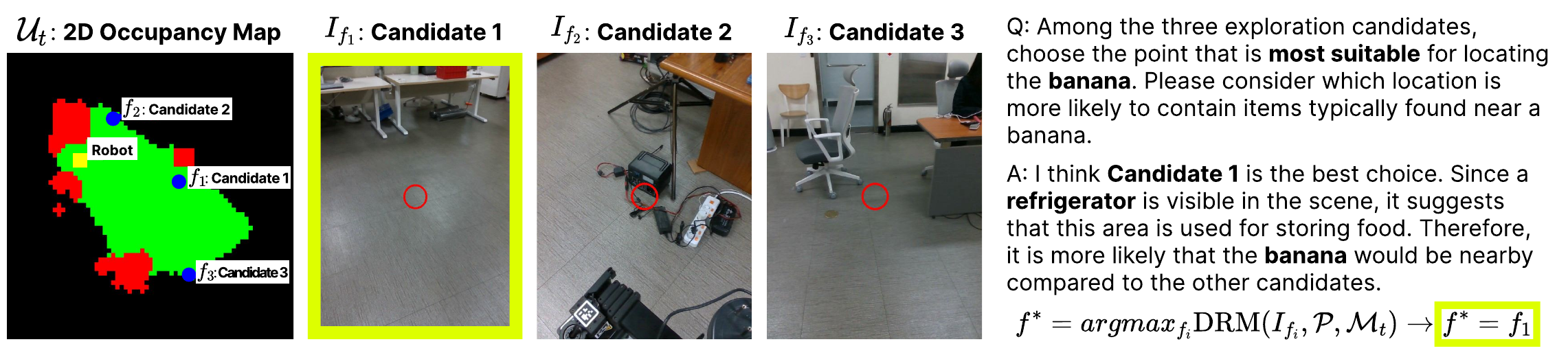}
    \caption{\noindent \textbf{DRM-based frontier selection with top-$k$ candidate evaluation.} 
    The robot identifies top-$k$ frontier candidates $\{f_1, f_2, f_3\}$ from the exploration value map $V_i = V_i^T + V_i^S$, and obtains the corresponding camera views $I_{f_i}$ by orienting the camera toward each candidate. 
    Given these views, the DRM evaluates $\text{DRM}(I_{f_i}, \mathcal{P}, \mathcal{M}_t)$ to determine which frontier is most promising for locating the target object; in this example, the DRM selects $f_1$ because the scene context (e.g., a refrigerator) suggests a higher likelihood of finding a \textit{banana} nearby.}
    \label{fig:select_k}
\end{figure*}

\vspace{0.5em}
\noindent \textbf{DRM.}
To implement the planning component of this coordination, we employ a multimodal LLM as the DRM, which operates at navigation targets or when triggered by the IRM.
Upon activation, the robot executes a \texttt{look\_around} primitive to capture surrounding views and performs 3D semantic scene reconstruction following prior work~\cite{liu2024dynamem}.
This reconstruction updates the memory $\mathcal{M}_t$ that maintains: 
(1) a 3D semantic scene representation $\mathcal{S}_t$, \ie, 3D voxel map, 
(2) a 2D occupancy projection $\mathcal{U}_t$ derived from $\mathcal{S}_t$ for effective spatial reasoning, encoding navigable areas, obstacles, and semantic labels,
(3) action history $\mathcal{H}_t = \{a_1,...,a_t\}$, and 
(4) an object registry $\mathcal{R}_t = \{(c_i, p_i)\}_{i=1}^{N_t}$ accumulating $N_t$ discovered objects with category $c_i$ and position $p_i=(x_i, y_i, z_i)$.
Using the task instruction $\mathcal{P}$ and memory $\mathcal{M}_t$, the DRM generates planning decisions:
\begin{equation}
a_{t+1}, o_{t+1}, \mathcal{G}_{t+1} = \text{DRM}(\mathcal{P}, \mathcal{M}_t)
\end{equation}
This yields three outputs: 
(1) next action $a_{t+1} \in \{\texttt{go\_to}, \texttt{explore}, \texttt{grasp}, \texttt{place}\}$, 
(2) target specification $o_{t+1}$ (coordinates for \texttt{go\_to}, locations for \texttt{explore}, or object/receptacle IDs for manipulation), and 
(3) task-specific guidance prompt $\mathcal{G}_{t+1}$ that refocuses the IRM's attention for the upcoming phase.
The robot's controller then executes the action–target pair $(a_{t+1}, o_{t+1})$, and the IRM is reinitialized with the updated guidance $\mathcal{G}_{t+1}$ for continuous background processing during execution.

\vspace{0.5em}
\noindent \textbf{IRM.}
For continuous perception during task execution, we employ a Video-LLM~\cite{qwen25vl2025} as the IRM, enabling continuous environmental monitoring throughout navigation and manipulation.
The Video-LLM processes video clips $v_t$ (recent frames from the continuous stream) with task-specific guidance prompt $\mathcal{G}_{t}$ provided by the DRM to generate a structured language report $Z_t$ that describes detected objects, task progress, and potential issues:
\begin{equation}
Z_t = \text{Video-LLM}(v_t, \mathcal{G}_{t}).
\end{equation}

Since the Video-LLM generates free-form language outputs whose structure varies with task context, we employ a guidance-conditioned parsing module $\Phi_{\mathcal{G}_t}$ (detailed procedures are in Sec.~\ref{subsec:task_execution}) to extract actionable information:
\begin{equation}
\Phi_{\mathcal{G}_t}:\; Z_t \;\mapsto\; (\mathcal{X}_t,\, m_t),
\end{equation}
where detected object information $\mathcal{X}_t$ contains object category and position pairs $(c_i, p_i)$ used to update the object registry $\mathcal{R}_t$, and execution mode $m_t \in \{\textsc{continue}, \textsc{adjust}, \textsc{replan}\}$ specifies the appropriate robot behavior.
Here, $\mathcal{G}_t$ is a text prompt from the DRM that reconfigures $\Phi_{\mathcal{G}_t}$'s extraction target: during navigation it instructs $\Phi_{\mathcal{G}_t}$ to match object mentions in $Z_t$ against task targets via embedding similarity to populate $\mathcal{X}_t$; during manipulation it instead shifts to assessing grasp quality and object stability cues to determine $m_t$.

The execution mode $m_t$ enables three levels of adaptation: (i) \textsc{continue} maintains current execution when no issues are detected, (ii) \textsc{adjust} applies immediate corrections for minor deviations (\eg, grasp refinement), and (iii) \textsc{replan} triggers DRM invocation for 3D semantic scene reconstruction and strategy revision when crucial environmental changes occur (\eg, target object appearing unexpectedly).
This design ensures the IRM functions as an effective continuous monitor—detecting opportunities and threats between discrete 3D updates—while maintaining computational efficiency through selective DRM activation.

\subsection{Task Execution Strategies}
\label{subsec:task_execution}

\noindent \textbf{Exploration and navigation.}
Our exploration strategy builds upon the value-guided frontier selection method from DynaMem~\cite{liu2024dynamem}, 
which combines temporal and semantic value maps to compute exploration values $V_i = V^T_i + V^S_i$, where $V^T_i$ prioritizes least-recently-seen areas and $V^S_i$ measures semantic similarity to target objects.
To compute the semantic value $V^S_i$, we follow the same open-vocabulary embedding mechanism used in DynaMem~\cite{liu2024dynamem}. Each frontier cell $i$ is associated with its most recent image, from which we extract an embedding using the CLIP image encoder. Similarly, the target object query is converted into a text embedding using the CLIP text encoder. The semantic similarity is then measured by the dot product between these embeddings. Frontiers whose appearance aligns more closely with the target text query therefore receive higher $V^S_i$.

We enhance this approach through DRM-based intelligent selection, as pure value-based ranking may overlook contextual cues visible from the current position.
To obtain a compact yet diverse set of candidate frontiers for DRM evaluation, we avoid directly taking the top elements of $V$, since high-value frontiers often cluster spatially and offer nearly identical viewpoints.
Instead, we sort all frontier cells by $V_i$ in descending order and iteratively add a frontier to the candidate set only when it lies beyond a fixed spatial threshold from those already selected.
This procedure suppresses redundant neighbors and yields a more informative, spatially distributed candidate set; in practice, it consistently produces three useful candidates ($k=3$) that balance coverage and computational cost. 
As illustrated in Fig.~\ref{fig:select_k}, the robot obtains the corresponding view $I_f$ for each selected frontier by orienting the camera toward it, enabling the DRM to directly compare the contextual relevance of each candidate.
Specifically, the DRM evaluates these top-$k$ candidates:
\begin{equation}
f^* = \operatorname*{arg\,max}_{f \in \text{top-k}(V)} \text{DRM}(I_f,\mathcal{P},\mathcal{M}_t)
\end{equation}
where $I_f$ denotes the image associated with frontier $f$.
This enables the DRM to leverage visual context alongside task instruction $\mathcal{P}$, memory $\mathcal{M}_t$ for context-aware destination selection.
Once $f^*$ is determined, the robot generates a trajectory using A* path planning~\cite{hart1968astar} and begins navigation.

During transit, the IRM operates in the background using the guidance prompt $\mathcal{G}_t$ provided by the DRM, processing short video clips and producing structured text reports $Z_t$ that describe scene elements relevant to the current phase.
The guidance-conditioned parsing module $\Phi_{\mathcal{G}_t}$ extracts actionable information from these free-form text reports, identifying both the set of detected object information $\mathcal{X}_t$ relevant to the task and the execution mode $m_t$.
Detected objects are identified by matching nouns in $Z_t$ with task-relevant entities specified in $\mathcal{P}$ using embedding similarity~\cite{song2020mpnet}; when a match is found, the system initiates an asynchronous localization step to estimate the corresponding 3D position.
OWL-ViT~\cite{minderer2022simple} detects the 2D bounding box $B_i$ of each matched object $i$, which is then lifted to 3D position $p_i$ using RGB-D projection:
\begin{equation}
    p_i = \text{median}\{T^{\text{cam}\rightarrow\text{world}}_t(u,v,d(u,v)) : (u,v) \in B_i\},
\end{equation}
where $T^{\text{cam}\rightarrow\text{world}}_t$ denotes the standard camera-to-world transformation using RGB-D measurements and camera parameters at time $t$, following~\cite{cheng2021back}.
The median operation within $B_i$ improves robustness to depth noise and background pixels.
The resulting positions form $\mathcal{X}_t$ and are merged into $\mathcal{R}_{t+1}$ without interrupting motion.
The execution mode $m_t$ is determined using the same guidance context; for example, if the IRM’s report indicates that a task-relevant object has newly appeared or that a navigation target has shifted, $\Phi_{\mathcal{G}_t}$ returns a \textsc{replan} signal, whereas otherwise it returns \textsc{continue} and updates the memory with $\mathcal{X}_t$ as appropriate.
This mechanism allows the IRM to supply timely and task-aware feedback without interrupting motion unless necessary.


\begin{figure}[h]
    \centering
    \includegraphics[width=0.9\linewidth]{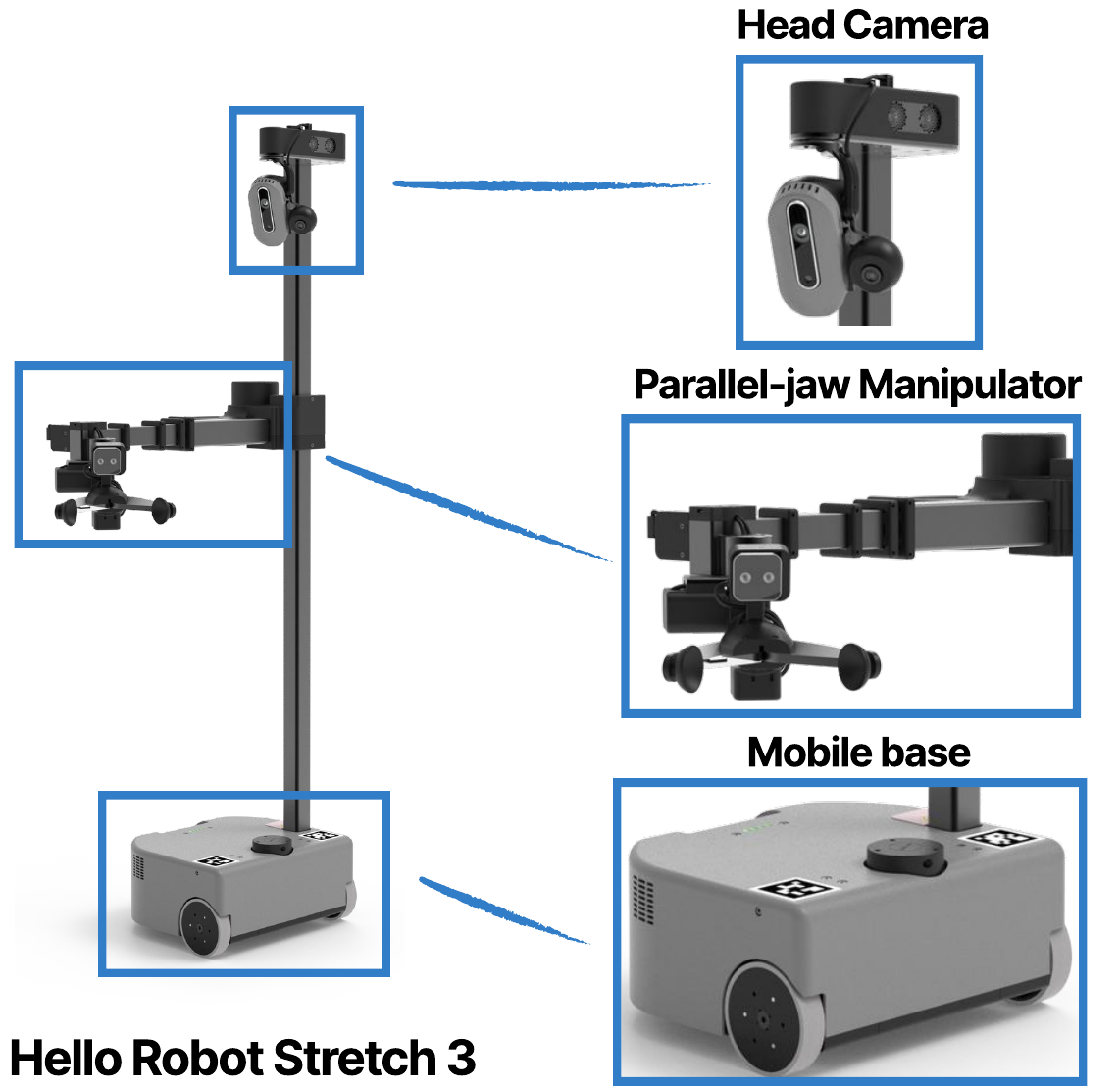}
    \caption{\textbf{Hello Robot Stretch SE3 used in our experiments.} Equipped with a mobile base, prismatic lift, 3-DoF wrist, and parallel-jaw gripper, the robot uses a head-mounted RealSense D435i for wide-view RGB-D observations (for exploration and 3D reconstruction) and a wrist-mounted RealSense D405 for accurate short-range depth during grasping. Low-level control and sensor streaming run on the onboard computer, while all LLM components (DRM and IRM) run on an external workstation over Wi-Fi; grasp poses generated by AnyGrasp are transformed into the robot frame and executed using Stretch’s inverse kinematics.}
    \label{fig:stretch}
\end{figure}

\begin{figure*}[t]
    \vspace{2mm}
    \centering
    \includegraphics[width=1\linewidth]{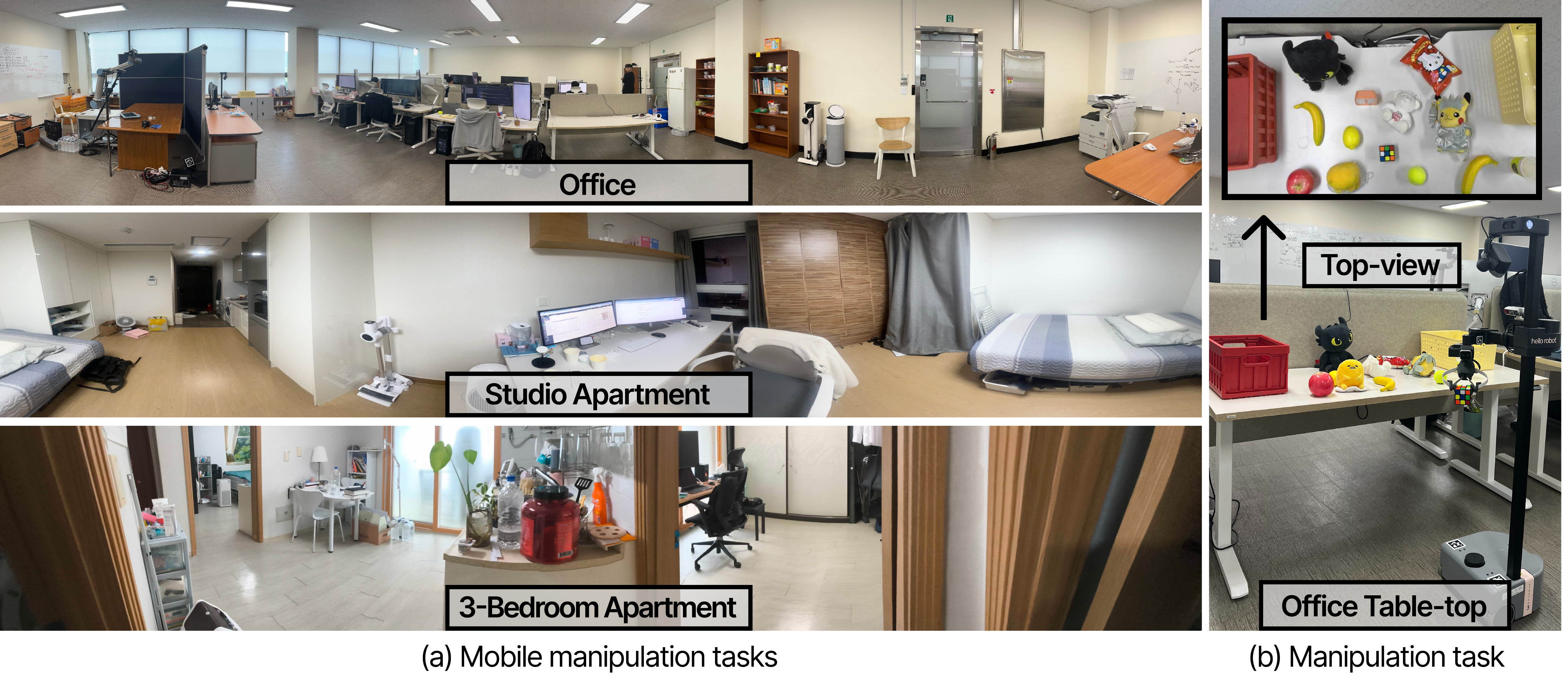}
    \caption{\noindent \textbf{Experimental environments.}
    (a) \textbf{Mobile manipulation:} We evaluate \method\ in a controlled office and two real-world home sites (a one-room studio and a three-room apartment) with varying layout complexity, where objects and receptacles form diverse multi-step OVMM scenes under identical code and configurations across all environments.
    (b) \textbf{Tabletop manipulation:} The robot's motion is limited to forward–backward translation; three objects and three receptacles are placed on a table, and we run 30 trials per condition to isolate the IRM’s contribution to manipulation performance.}
    \label{fig:envs}
\end{figure*}

\vspace{0.5em}
\noindent \textbf{Manipulation.}
Our manipulation approach builds on OK-Robot~\cite{liu2024okrobot}, which combines AnyGrasp~\cite{fang2023anygrasp} with LangSAM~\cite{medeiros2023langsam} filtering for grasping and uses point cloud-based height computation for placing. 
We extend this framework with event-triggered visual feedback through the IRM, enabling reactive adjustments in dynamic environments.

During manipulation, $\mathcal{X}_t$ typically remains empty as objects are already localized, while $\Phi_{\mathcal{G}_t}$ focuses on extracting the execution mode $m_t$ from the IRM's reports $Z_t$.
Similar to the navigation phase, $\Phi_{\mathcal{G}_t}$ analyzes $Z_t$ using embedding similarity~\cite{song2020mpnet} to identify manipulation-specific cues such as grasp quality indicators, object stability assessments, and environmental changes.
When the IRM detects misalignments during grasping ($m_t$ = \textsc{adjust}), we perform local grasp recomputation: AnyGrasp generates new candidates within a constrained region around the current target, selecting the highest-scoring pose with minimal reorientation.
This enables immediate corrections without costly full replanning overhead.
For placing, the IRM monitors object stability and receptacle availability continuously throughout execution.
When issues arise, $\Phi_{\mathcal{G}_t}$ returns $m_t$ = \textsc{adjust}, triggering height recomputation or alternative receptacle selection based on the problem detected.
Critical failures---such as repeated grasp failures or unavailable receptacles---result in $m_t$ = \textsc{replan}, engaging the DRM for strategic revision.
The DRM then performs updated 3D reconstruction and generates alternative strategies, such as selecting different objects or modifying task sequences.

This interaction between the DRM and IRM makes the overall manipulation process substantially more robust than open-loop execution, as adjustments are applied as soon as issues arise rather than after a failure has already occurred.
It is also more efficient than approaches that evaluate success only after an entire manipulation attempt is finished and then restart from the beginning upon failure, since IRM-guided corrections allow the robot to recover mid-execution without discarding prior progress.

\section{Experiments}
We evaluate \method in real-world environments to assess its robustness against environmental changes introduced during task execution and its effectiveness on long-horizon multi-object tasks compared to baselines.

\subsection{Experimental Settings}

\noindent\noindent \textbf{Robot setups.}
We use a Hello Robot Stretch SE3~\cite{hellorobot2023stretch} equipped with a mobile base, prismatic lift, 3-DoF wrist, and parallel-jaw gripper, as shown in Fig.~\ref{fig:stretch}.
For perception, the robot uses a head-mounted RealSense D435i for wide-view RGB-D observations (\eg, exploration and 3D reconstruction) and a wrist-mounted RealSense D405 for accurate short-range depth during grasping.
Low-level control and sensor streaming run on the onboard computer, while all LLM components (DRM and IRM) run on an external workstation over Wi-Fi.
Grasp poses generated by AnyGrasp are transformed into the robot frame and executed using Stretch’s inverse kinematics.

\vspace{1em}
\noindent\textbf{Implementation details.}
Our system builds upon DynaMem's 3D voxel representation~\cite{liu2024dynamem}.
We employ GPT-5~\cite{openai2025gpt5systemcard} as our DRM and Qwen2.5VL (3B)~\cite{qwen25vl2025} as our Video-LLM.
The Video-LLM processes 1-second clips at 8 fps with an inference time of about 0.5 seconds per clip, requiring approximately 12\,GB of VRAM on a single NVIDIA A6000 (48\,GB) without interfering with the navigation, perception, or control pipelines sharing the same GPU. This throughput is sufficient to keep up with the robot's motion during task execution, allowing the IRM to provide frequent updates without delaying control actions.

\subsection{Task Setup}

\noindent \textbf{Multi-step tasks in dynamic environments.}
Following previous work~\cite{comerobot2024}, we systematically evaluate multi-step task execution by defining three task categories with increasing complexity:
\textbf{Task 1:} Single object $\rightarrow$ single receptacle.
\textbf{Task 2:} Two objects $\rightarrow$ two receptacles.
\textbf{Task 3:} Three objects $\rightarrow$ three receptacles.
Experiments are conducted in three environments: a controlled office, a studio apartment, and a three-room apartment (Fig.~\ref{fig:envs}-(a)), covering both structured and more cluttered real-world settings under a unified protocol.
We evaluate all three task categories in the office (40 trials each) and focus on Task 3 in the homes (10 trials each), using identical code across all settings so that differences in performance can be attributed to the environment and task complexity rather than implementation details.

We vary three key factors:
(1) \textbf{Scenes}: Each unique object–receptacle arrangement defines a distinct initial state, which is maintained consistently across all compared methods to enable fair, scene-wise comparison.
(2) \textbf{Queries}: Task instructions specify randomly sampled object–receptacle pairs (1–3 pairs based on task category), ensuring that the same high-level queries are issued to every method.
(3) \textbf{Dynamics}: We introduce two position perturbations per query—typically moving objects during approach and receptacles during transport—simulating real-world dynamics where both targets and receptacles may shift while the robot is executing action.

\begin{table*}[t]
\centering
\small
\caption{\noindent \textbf{Real-world office environment evaluation across Task~1--3.}
The three task categories contain 1, 2, and 3 object$\rightarrow$receptacle subtasks, respectively, testing increasing difficulty from single-step to long-horizon execution.
We report four metrics: SR for overall task completion, PSR for subgoal progress, SPL for path efficiency, and PSPL for efficiency on partially completed tasks (for Task~1, SPL and PSPL are equivalent).
Across all three tasks, \method consistently achieves the highest SR, PSR, SPL, and PSPL, and maintains strong performance even as the number of subtasks increases, whereas baseline methods degrade sharply with task complexity.}
\label{tab:lab_main}
\resizebox{0.9\linewidth}{!}{%
\begin{tabular}{lcc cccc cccc}
\toprule
\multirow{2}{*}{\noindent \textbf{Method}} 
& \multicolumn{2}{c}{\noindent \textbf{Task 1 (1 subtask)}} 
& \multicolumn{4}{c}{\noindent \textbf{Task 2 (2 subtasks)}} 
& \multicolumn{4}{c}{\noindent \textbf{Task 3 (3 subtasks)}} \\
\cmidrule(lr){2-3} \cmidrule(lr){4-7} \cmidrule(lr){8-11}
& \noindent \textbf{SR} $\uparrow$ 
& \noindent \textbf{SPL} $\uparrow$ 
& \noindent \textbf{SR} $\uparrow$ 
& \noindent \textbf{PSR} $\uparrow$ 
& \noindent \textbf{SPL} $\uparrow$ 
& \noindent \textbf{PSPL} $\uparrow$
& \noindent \textbf{SR} $\uparrow$ 
& \noindent \textbf{PSR} $\uparrow$ 
& \noindent \textbf{SPL} $\uparrow$ 
& \noindent \textbf{PSPL} $\uparrow$ \\
\midrule
OK-Robot &  0.23  &  0.20 &   0.05 &  0.19  &  0.05  &  0.19  &  0.03  &  0.27  &  0.03  &  0.13 \\
DovSG    &  0.28  &  0.25 &  0.13  &  0.23  &  0.13  &  0.16  &  0.08  &  0.36  &  0.05 &  0.23  \\
DynaMem  &  0.60  &  0.42  &  0.43  &  0.71  &  0.29  &  0.40  &  0.15  &  0.62  &  0.09  &  0.47  \\
\rowcolor{gray!10}
\method~(\noindent \textbf{Ours}) 
        & \textbf{0.93} & \textbf{0.69} & \textbf{0.78} & \textbf{0.88} & \textbf{0.68} & \textbf{0.71} & \textbf{0.63} & \textbf{0.85} & \textbf{0.48} & \textbf{0.72} \\
\bottomrule
\end{tabular}}
\end{table*}

\vspace{1em}
\noindent \textbf{Metrics.}
Following prior work~\cite{ALFRED20}, we use three standard metrics to quantify performance.
\textit{Success Rate} (SR) measures full task completion, \ie, whether all required high-level goals for a given instruction are satisfied.
\textit{Partial Success Rate} (PSR) instead averages the fraction of completed subgoals in multi-step tasks, capturing cases where the agent makes partial but meaningful progress even if the overall task is not fully completed.
\textit{Success weighted by Path Length} (SPL) combines task success with path efficiency, normalizing the executed trajectory length by expert demonstrations obtained from voxel-derived occupancy grids so that shorter, more efficient paths are rewarded.
For multi-subgoal tasks, we additionally introduce \textit{Partial Success weighted by Path Length} (PSPL), which extends SPL by averaging path efficiency over individual completed subgoals rather than requiring full task completion. 
PSPL mirrors the notion of PSR in the path-efficiency space, providing a more fine-grained view of how efficiently each successfully completed subgoal is achieved.

\vspace{1em}
\noindent \textbf{Baselines.}
We compare \method with three strong baselines for the OVMM task: OK-Robot~\cite{liu2024okrobot}, DynaMem~\cite{liu2024dynamem}, and DovSG~\cite{yan2024dovsg}.
Since OK-Robot and DynaMem are originally designed for single object–receptacle tasks, we extend them to our multi-object setting by sequentially executing each object–receptacle pair specified in the instruction query, treating the multi-object instruction as an ordered list of independent subtasks.
Both OK-Robot and DovSG require a global pre-scanning phase to build environment maps before any manipulation begins, and their performance is highly sensitive to the quality of this initial scan.
To ensure a fair comparison under this sensitivity, we perform five separate scans per scene and report results using the best-quality map obtained in that scene.
For DovSG, we additionally replace the Stretch SE3's default D435i camera with a RealSense D455 RGB-D camera, following their original implementation, as ACE-based pose estimation was unreliable with the default hardware.

\begin{table}[t]
  \centering
  \caption{\textbf{Real-world home environment evaluation on Task~3.} Results are shown for a studio apartment and a three-bedroom apartment. We report Success Rate (SR) and Success weighted by Path Length (SPL). Across both homes, \method achieves substantially higher SR than the strongest baseline (DynaMem), with improvements of roughly 0.4 in the studio and 0.3 in the three-room layout, while SPL nearly triples and doubles, respectively, indicating that our dual-process design yields not only more successful but also more path-efficient executions in dynamic home environments.}
  \label{tab:home_main}
  \resizebox{0.79\linewidth}{!}{%
  \begin{tabular}{lcccc}
    \toprule
    \multirow{2}{*}{\textbf{Method}} & \multicolumn{2}{c}{\textbf{Studio}} & \multicolumn{2}{c}{\textbf{3-Room}} \\
    \cmidrule(lr){2-3}\cmidrule(lr){4-5}
     & \textbf{SR} & \textbf{SPL} & \textbf{SR} & \textbf{SPL} \\
    \midrule
    OK\text-Robot & 0.20 & 0.10 & 0.20 & 0.18 \\
    DovSG         & 0.20 & 0.20 & 0.40 & 0.33 \\
    DynaMem       & 0.30 & 0.15 & 0.50 & 0.36 \\
    \rowcolor{gray!10}\method~(\textbf{Ours}) & \textbf{0.70}  & \textbf{0.57} & \textbf{0.80} & \textbf{0.62} \\
    \bottomrule
  \end{tabular}}
\end{table}

\subsection{Quantitative Results}

\noindent \textbf{Quantitative results in office environment.}
Table~\ref{tab:lab_main} demonstrates \method's consistent superiority across all task complexities. In single-object tasks (Task~1), \method achieves an SR of 0.93 compared to 0.60 for the best baseline (DynaMem), already indicating a large gap even in the simplest scenario. 
This advantage amplifies as task complexity increases: in Task~2 (two subtasks), \method reaches 0.78 versus DynaMem's 0.43, and in Task~3 (three subtasks), it still maintains an SR of 0.63---over 4$\times$ higher than any baseline, despite the increased risk of compounding failures across multiple subtasks. 
Moreover, \method excels in partial task completion, achieving a PSR of 0.85 in Task~3 compared to DynaMem's 0.62, indicating robust recovery from individual failures and the ability to complete remaining subgoals even when some attempts do not succeed.

These improvements stem from our dual-process design: the IRM enables event-triggered grasp corrections and dynamic adjustments through continuous monitoring, while the DRM ensures efficient exploration via top-$k$ frontier evaluation (Sec.~\ref{subsec:task_execution}), so that the system can both react locally and plan globally within the same execution. 
This is reflected in improved SPL/PSPL metrics and shorter trajectories (Table~\ref{tab:time_path}), showing that higher success does not come at the cost of longer or less efficient paths. 
Our heterogeneous compute strategy effectively resolves the fundamental trade-off in existing approaches---OK-Robot and DovSG suffer from stale perception, while DynaMem incurs costly reconstruction pauses. 
By decoupling strategic planning (DRM) from lightweight monitoring (IRM), \method achieves both temporal continuity and spatial precision, and this combination directly translates to higher success rates and more efficient task execution under dynamic conditions.



\vspace{0.5em}
\noindent \textbf{Quantitative results in home environments.}
Table~\ref{tab:home_main} shows \method's clear advantages in both home settings, with success rates improving by roughly 0.4 in the one-room studio and 0.3 in the three-room layout compared to DynaMem, indicating a substantial margin over the strongest baseline in these scenarios. 
These gains stem from the IRM's opportunistic target detection during navigation, which allows the system to exploit newly observed opportunities as the robot moves, and from the DRM's timely replanning in response to changed receptacle states when the environment does not remain static.

Efficiency metrics mirror these improvements as well: SPL improves by ${\sim}3.8{\times}$ in the studio and ${\sim}1.7{\times}$ in the 3-room apartment, closely tracking the boost in success rates and confirming that trajectories also become more economical. 
Taken together, these results confirm that our dual-process design enhances both reliability and path efficiency under dynamic conditions, yielding executions that are not only more successful but also more efficient in path usage.

\begin{table}[t]
\centering
\caption{\noindent \textbf{Ablation study of proposed dual-process components.} 
We evaluate four variants of \method. 
Experiments are conducted on Task~3 (three objects $\rightarrow$ three receptacles) in the office environment with 10 trials per variant. 
Metrics include SR, PSR, and SPL.}
\label{tab:ablation}
\resizebox{\linewidth}{!}{%
\begin{tabular}{lcccccc}
\toprule
\multirow{2}{*}{\textbf{Configuration}} & \multicolumn{2}{c}{\textbf{Components}} & \multirow{2}{*}{\textbf{SR}} & \multirow{2}{*}{\textbf{PSR}} & \multirow{2}{*}{\textbf{SPL}} & \multirow{2}{*}{\textbf{PSPL}} \\
\cmidrule(lr){2-3}
 & \textbf{DRM} & \textbf{IRM} &  &  &  \\
\midrule
Neither   & \color[HTML]{CB0000}{\ding{56}} & \color[HTML]{CB0000}{\ding{56}} & 0.30 & 0.43 & 0.22 & 0.28 \\
DRM only            & \color[HTML]{036400}{\ding{52}} & \color[HTML]{CB0000}{\ding{56}} & 0.40 & 0.57 & 0.38 & 0.50 \\
IRM only            & \color[HTML]{CB0000}{\ding{56}} & \color[HTML]{036400}{\ding{52}} & 0.60 & 0.83 & 0.47 & 0.59 \\
\rowcolor{gray!10}
    DRM+IRM (\method) & \color[HTML]{036400}{\ding{52}} & \color[HTML]{036400}{\ding{52}} & \textbf{0.80} & \textbf{0.93} & \textbf{0.63} & \textbf{0.82} \\\bottomrule
\end{tabular}%
}
\end{table}

\subsection{In-depth analysis}



\noindent \textbf{Ablation study.}
To assess the contribution of the Deliberative Reasoning Module (DRM) and the Instant Response Module (IRM), we evaluate four variants on Task~3 with 10 trials each, as illustrated in Table~\ref{tab:ablation}:
\textit{DRM + IRM} (full system with both modules enabled),
\textit{DRM only} (deliberative planning with contextual frontier selection but without continuous monitoring),
\textit{IRM only} (continuous monitoring without DRM guidance), and
\textit{Neither} (both modules disabled, relying only on discrete voxel-map updates).

\textit{DRM only} delivers modest gains in SR/PSR (SR: $0.30\!\rightarrow\!0.40$, PSR: $0.43\!\rightarrow\!0.57$), but achieves more substantial improvements in path efficiency (SPL: $0.22\!\rightarrow\!0.38$, PSPL: $0.28\!\rightarrow\!0.50$). This improvement stems from the DRM’s contextual frontier evaluation which considers top-$k$ candidates, and avoids unnecessary detours during exploration.
\textit{IRM only}, run without DRM guidance, relies on generic scene descriptions from the Video-LLM rather than task-conditioned monitoring. 
This configuration improves reliability by providing continuous perception and enabling failure recovery(SR: $0.30\!\rightarrow\!0.60$, PSR: $0.43\!\rightarrow\!0.83$). 
However, the absence of DRM guidance limits its effectiveness, since the IRM cannot prioritize what aspects of the scene to attend to during action execution.
Finally, the combined \textit{DRM + IRM} system achieves the strongest synergy: SR rises further to $0.80$ and SPL to $0.63$, with corresponding PSR $0.93$ and PSPL $0.82$. 
This confirms the dual-process design from Sec.~\ref{sec:intro}: while the IRM ensures temporal continuity through opportunistic detections and micro-corrections, the DRM provides task-aware guidance that tells the IRM where to focus during execution and supplies timely replanning when needed. 
This combination balances temporal awareness with spatial precision, resulting in more reliable and efficient task execution than either module can provide on its own.

\begin{table}[t]
\centering
\caption{\noindent \textbf{Effect of IRM on tabletop manipulation tasks.} 
We compare a baseline without IRM (DRM only) against our system with IRM enabled (DRM + IRM, \ie, \method), averaging results over 30 manipulation trials with restricted base motion. 
Metrics are overall success rate and average execution time, highlighting how IRM improves reliability with minimal time overhead.}
\label{tab:manip_ablation_dual}
\resizebox{0.95\linewidth}{!}{
\begin{tabular}{lcc}
\toprule
\noindent \textbf{Configuration} & \noindent \textbf{SR} $\uparrow$ & \noindent \textbf{Avg. Time (sec.)} $\downarrow$ \\
\midrule
DRM only & 0.53 & \textbf{61} \\
DRM + IRM (\method)    & \textbf{0.77} & 66 \\
\bottomrule
\end{tabular}
}
\end{table}

\begin{figure*}[!t]
  \centering
  \includegraphics[width=1\linewidth]{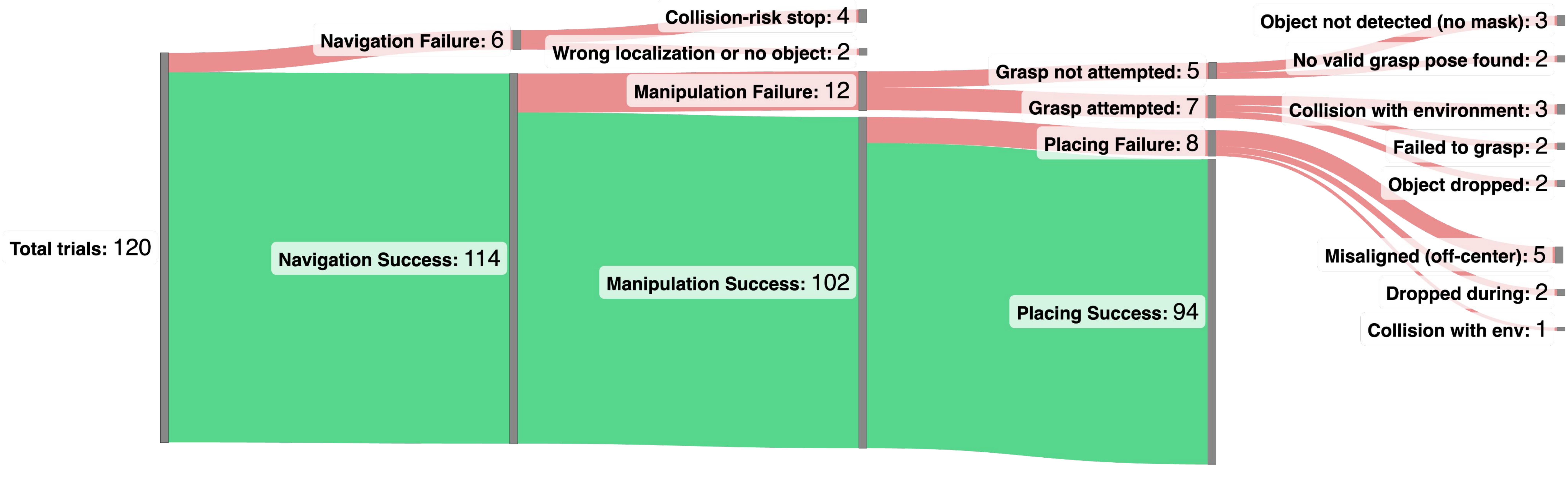}
  \caption{\noindent \textbf{Failure analysis of \method across Task~1--3.}
    Sankey diagram over 120 trials (40 per task) showing how episodes progress through navigation, manipulation, and placing. From 120 trials, 114 reach the target region (navigation success) while 6 fail, mostly due to conservative collision-risk stops (4) and wrong localization or missing objects (2). Among the 114 successful navigations, 102 complete the grasp (manipulation success) and 12 fail: 5 cases where grasping is not attempted because the object is not detected (3) or no valid grasp pose is found (2), and 7 cases where an attempted grasp fails due to collisions with the environment (3), failed grasps (2), or dropped objects (2). Finally, of the 102 successful grasps, 94 complete the placing step, with 8 placing failures arising from misaligned (off-center) placements (5), objects dropped during placing (2), or collision with the environment (1). The figure highlights that remaining errors are concentrated in fine-grained perception and contact handling rather than high-level navigation.}
  \label{fig:failure_sankey}
\end{figure*}

\vspace{0.5em}
\noindent \textbf{Effect of IRM on manipulation.}
To further evaluate the contribution of the IRM, we conduct a tabletop study in a more constrained setting, where the robot base is restricted to only forward–backward motion, as shown in Fig.~\ref{fig:envs}-(b). 
In this setup, we place a diverse set of objects and receptacles on the table and run 30 manipulation trials with randomly sampled object–receptacle pairs under two configurations: \textit{with IRM enabled} \vs \textit{without IRM}. 
As reported in Table~\ref{tab:manip_ablation_dual}, enabling the IRM raises the success rate from 0.53 to 0.77, while incurring only a small increase in average execution time (61s $\rightarrow$ 66s), most of which is attributable to the few extra seconds spent re-aligning the gripper during local adjustments. 

The reliability gains stem from the IRM’s continuous monitoring of the manipulation process, which detects minor pose errors or object shifts as they occur and issues immediate local adjustments that refine the ongoing action, thereby correcting these deviations without requiring full replanning.

\begin{table}[th]
\centering
\small
\caption{\textbf{Comparison of completion time and trajectory length for Task~3 in the office environment.}
We report average completion time (minutes) and path length (meters), computed over successful trials from 40 runs.
Results compare our method against DynaMem~\cite{liu2024dynamem} and show that, despite using additional IRM/DRM modules, \method achieves both faster completion and shorter trajectories, consistent with the efficiency gains observed in our results.}
\label{tab:time_path}
\resizebox{\linewidth}{!}{
\begin{tabular}{lcc}
\toprule
\textbf{Method} & \textbf{Avg. Time (min.)} $\downarrow$ & \textbf{Avg. Length (m)} $\downarrow$ \\
\midrule
DynaMem~\cite{liu2024dynamem} &  33.83  &  39.60  \\
\rowcolor{gray!10}
\method &  \textbf{21.90}  &  \textbf{28.35}  \\
\bottomrule
\end{tabular}}
\end{table}

\vspace{0.5em}
\noindent \textbf{Completion time and path efficiency.}
Table~\ref{tab:time_path} shows that, despite the presence of additional modules (IRM and DRM), \method still achieves faster overall execution and produces shorter trajectories than DynaMem~\cite{liu2024dynamem}, which is the baseline with the highest SR in the office environment. 
This observation is consistent with our motivation in Sec.~\ref{sec:intro}: whereas prior systems repeatedly pause their motion to perform explicit map updates, \method relies on the IRM to provide continuous monitoring of the scene and to trigger updates only when the robot is actually exploring new areas. 

In this way, object localization is handled directly through the IRM during navigation, instead of requiring separate stopping phases, which results in smoother and more streamlined execution. 
Consequently, DynaMem~\cite{liu2024dynamem} trajectories end up being approximately 1.4$\times$ longer, as the robot must travel additional distance before objects are recognized and can be acted upon.

\vspace{0.5em}
\noindent \textbf{Failure analysis.}
Figure~\ref{fig:failure_sankey} presents a failure analysis of our method over the 120 trials used for Task~1–3 evaluation (40 trials per task). 
Starting from all trials on the left, the first split shows that navigation is highly reliable: 114 of 120 episodes successfully reach the target region, while only 6 terminate in navigation failure. 
Among these 6 cases, 4 correspond to conservative collision-risk stops, where execution was manually halted just before a potential collision, and 2 are due to wrong localization or arriving at a location where the target object is not found.

Conditioned on successful navigation, 102 of the 114 trials complete the manipulation (grasping) phase, whereas 12 fail during manipulation. These 12 trials further divide into 5 cases where a grasp is never attempted and 7 where a grasp is attempted but unsuccessful. 
When grasping is not attempted, the failure is driven by perception or grasp-planning limitations: 3 trials where the object is not detected (no valid segmentation mask) and 2 where no valid grasp pose is found. 
For the 7 trials with an attempted grasp, failures arise from physical interaction issues—3 collisions with the environment, 2 failed grasps, and 2 instances where the object is initially picked up but then dropped.
Finally, among the 102 successful grasps, 94 episodes also succeed in placing the object in the receptacle, yielding an overall placing success of 94/120 trials. 
The remaining 8 placing failures are dominated by fine-grained pose and stability errors: 5 misaligned (off-center) placements, 2 drops during the placing motion, and 1 collision with the environment.

\begin{figure}[!t]
  \centering
  \includegraphics[width=1\linewidth]{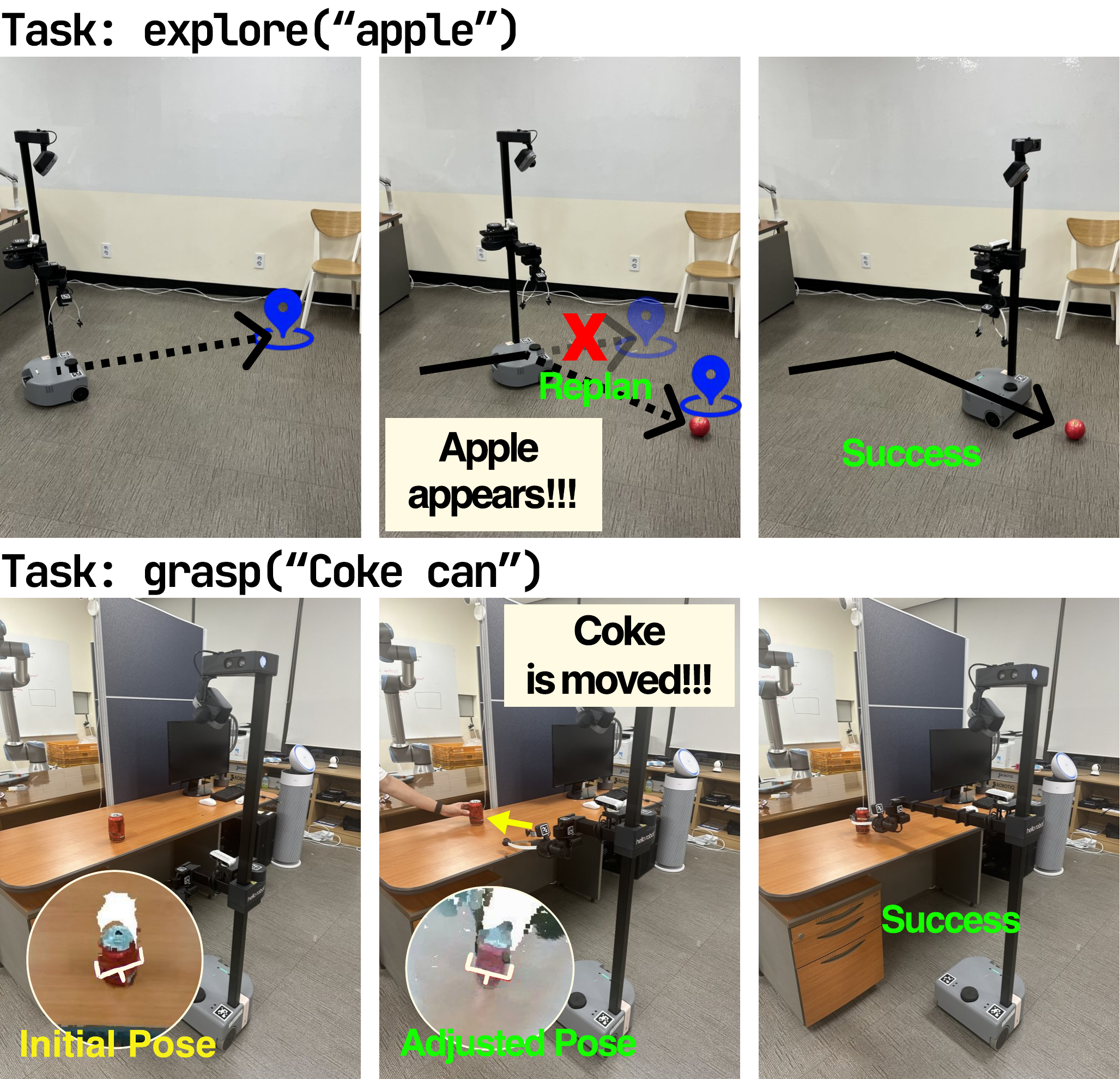}
  \caption{\noindent \textbf{Qualitative examples of \method in dynamic environments.}
    Top: \textit{Exploration.} An apple appears mid-navigation; the IRM detects it and triggers DRM replanning, leading to efficient target acquisition.
    Bottom: \textit{Manipulation.} A Coke can is displaced during grasp; the IRM detects the shift, adjusts the pose, and completes the action without full replanning. Together, DRM and IRM maintain temporal awareness and spatial precision under dynamic changes.}
  \label{fig:quali}
\end{figure}

\vspace{0.5em}
\subsection{Qualitative Results}
We illustrate in Fig.~\ref{fig:quali} how \method adapts to dynamic changes that occur during execution. 
In the top example in Fig.~\ref{fig:quali}, an apple appears mid-navigation; the IRM detects the newly visible object in the incoming video stream and immediately triggers a \textsc{replan}, allowing the DRM to update the current plan so that the robot can divert from its original route and grasp the object efficiently. 
In the bottom example, a Coke can is displaced during the grasping phase; in response, the IRM issues an \textsc{adjust} event, enabling rapid re-alignment of the gripper with the shifted object and successful completion of the grasp without restarting the manipulation pipeline from scratch. 
Taken together, these examples illustrate how continuous monitoring and deliberative replanning work in concert to preserve both temporal awareness of changes as they happen and spatial precision in object interactions during real-world execution.
\section{Conclusion}
We presented \method, a dual-process framework that addresses OVMM's core limitation---\emph{intermittent scene perception}---by decoupling continuous video monitoring (IRM) from selective 3D reconstruction and planning (DRM) via bidirectional coordination.
Across an office and two real-world homes, \method consistently improved metrics and reduced time and path length over baselines.
Ablations confirmed the roles of DRM and IRM, and tabletop studies showed higher manipulation reliability.
By maintaining continuous awareness between updates while preserving geometry-accurate planning at key decision points, \method advances OVMM toward robust real-world deployment.

\section*{Acknowledgement}
This work was partly supported by the IITP grants (RS-2022-II220077, RS-2022-II220113, RS-2022-II220959, RS-2022-II220871, RS-2021-II211343 (SNU AI), RS-2025-25442338 (AI Star Fellowship-SNU)) funded by the Korea government (MSIT), grants (RS-2025-25462891 (US-KOR BARI), RS-2025-25453780) funded by MOTIR, a grant of Korean ARPA-H Project through the Korea Health Industry Development Institute (KHIDI), funded by the Ministry of Health \& Welfare, Republic of Korea (RS-2025-25424639), and the BK21 FOUR program, SNU in 2025, and the Artificial Intelligence Industrial Convergence Cluster Development Project funded by the Ministry of Science and ICT (MSIT, Korea) \& Gwangju Metropolitan City.

\addtolength{\textheight}{-12cm}   








\bibliographystyle{IEEEtran}
\bibliography{IEEEabrv,IEEEexample}

\end{document}